\documentclass[nohyperref]{article}

\usepackage{microtype}
\usepackage{graphicx}
\usepackage[lofdepth, lotdepth]{subfig}
\usepackage{caption}
\usepackage{booktabs} 

\usepackage{hyperref}

\usepackage[accepted]{icml2022}

\usepackage{amsmath}
\usepackage{amssymb}
\usepackage{mathtools}
\usepackage{amsthm}
\usepackage[lofdepth, lotdepth]{subfig}

\usepackage[capitalize,noabbrev]{cleveref}

\theoremstyle{plain}

\theoremstyle{definition}

\theoremstyle{remark}

\usepackage[textsize=tiny]{todonotes}

\def\eg{\emph{e.g.}}

\icmltitlerunning{}

\begin{document}

\twocolumn[
\icmltitle{A Light-weight Interpretable Model for Nuclei Detection\\
and Weakly-supervised Segmentation}



\icmlsetsymbol{equal}{*}

\begin{icmlauthorlist}
\icmlauthor{Yixiao Zhang}{jhu}
\icmlauthor{Adam Kortylewski}{max}
\icmlauthor{Qing Liu}{jhu}
\icmlauthor{Seyoun Park}{jhu}
\icmlauthor{Benjamin Green}{jhu}
\icmlauthor{Elizabeth Engle}{jhu}
\icmlauthor{Guillermo Almodovar}{jhu}
\icmlauthor{Ryan Walk}{jhu}
\icmlauthor{Sigfredo Soto-Diaz}{jhu}
\icmlauthor{Janis Taube}{jhu}
\icmlauthor{Alex Szalay}{jhu}
\icmlauthor{Alan Yuille}{jhu}
\end{icmlauthorlist}

\icmlaffiliation{jhu}{Department of Computer Science, Johns Hopkins University, Baltimore, US}
\icmlaffiliation{max}{Max Planck Institute for Informatics, Saarbrücken, Germany}

\icmlcorrespondingauthor{Yixiao Zhang}{yzhan334@jhu.edu}

\icmlkeywords{Nuclei detection and segmentation, Compositional Networks}

\vskip 0.3in
]



\printAffiliationsAndNotice{}  

\begin{abstract}

The field of computational pathology has witnessed great advancements since deep neural networks have been widely applied. These networks usually require large numbers of annotated data to train vast parameters. However, it takes significant effort to annotate a large histopathology dataset. We introduce a light-weight and interpretable model for nuclei detection and weakly-supervised segmentation. It only requires annotations on isolated nucleus, rather than on all nuclei in the dataset. Besides, it is a generative compositional model that first locates parts of nucleus, then learns the spatial correlation of the parts to further locate the nucleus. This process brings interpretability in its prediction.
Empirical results on an in-house dataset show that in detection, the proposed method achieved comparable or better performance than its deep network counterparts, especially when the annotated data is limited. It also outperforms popular weakly-supervised segmentation methods. The proposed method could be an alternative solution for the data-hungry problem of deep learning methods.
\end{abstract}

\section{Introduction}
\label{introduction}

Histopathology images provide an understanding of the microenvironment of various diseases. Nuclei detection and segmentation plays an important role for the analysis of cell morphology and organization. Unfortunately, the non-uniform chromatin texture, irregularity in size and shape as well as touching cells and background clutters put a big challenge to automated nuclei detection and segmentation. Some recently proposed methods rely on symmetry and stability of cellular regions \cite{arteta2012learning,cosatto2008grading,kuse2011local,veta2013automatic}. However, these method mainly focus on the detection of regular shaped cells and suffer from atypical characteristics.

More recently, deep convolutional neural networks (DNNs) have shown remarkable and reliable performance in various medical image detection and segmentation including histopathology images \cite{Tofighi2018,Xu2016,Hofener2018,Naylor2017,Kumar2017,Graham2019}. \cite{liu2021mdc} proposed a multi-scale connected segmentation network with distance map and contour information for nuclei segmentation. \cite{chanchal2021high} proposed a network which used dense residual connections and an atrous spatial pyramid pooling (ASPP) layer to learn rich multi-scale features. \cite{gao2021nuclei} used a Composite High-resolution Network for the segmentation and grading of clear cell renal cell carcinoma.  However, the collection of a large number of annotated data is critical and becomes a bottleneck to train conventional DNNs for the analysis of new modalities \cite{Lee2020,Qu2020a}. In addition, most works regard DNNs as a black box discriminative model without exploring its intermediate representations, thus having little interpretability in their predictions. Considering cell structures, especially nuclei shapes are invariant by stains, generative models for nuclei detection and segmentation learned from a small dataset are an alternative for efficient and robust analysis of various histopathology images.

\begin{figure*}[ht]
\vskip 0.2in
\begin{center}
\includegraphics[width=0.9\textwidth]{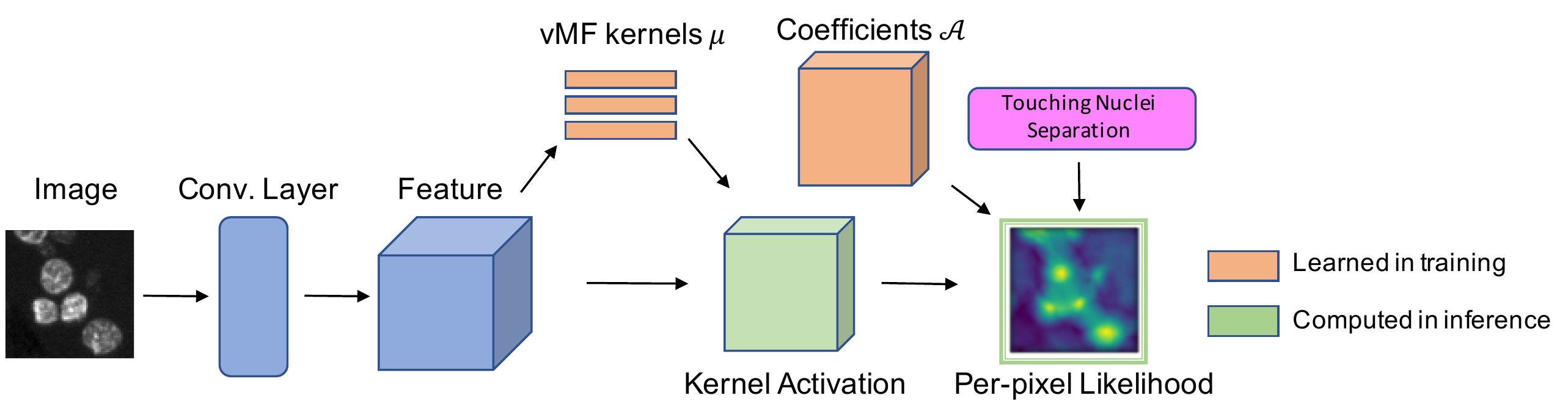}
\caption{Flowchart of the proposed method for nuclei detection. A convolution layer is used as feature extractor. For training, we cropped nucleus image patches and learn model parameters ($\mu$ and $\mathcal{A}$) in an unsupervised way. During testing, we compute the likelihood of existence of nucleus with learned parameters, together with a shape decomposition algorithm to separate touching nuclei, to obtain a final per-pixel likelihood prediction.}
\label{fig:flowchart}
\end{center}
\vskip -0.1in
\end{figure*}

In this study, we propose a light-weight interpretable model for nuclei detection and weakly supervised segmentation. We aim to design a generative model for a single nucleus, therefore only annotations on isolated nucleus are required to learn this model, which significantly reduces the annotation cost.
Inspired by the Compositional Networks \cite{Kortylewski2020}, we developed a compositional model that first find signals of parts of a nucleus, then spatially combine the signals of parts to determine the position of nucleus. In this way, the proposed method is able to locate nuclei by finding image regions that it can explain with high probability, while also provide human interpretable explanations for its prediction, due to the explicit compositional modeling of nuclei.
To the best of our knowledge, we are the first to adapt Compositional Networks to nuclei detection and segmentation on histopathology images.

To further boost the performance at touching cells that are hard to locate and segment precisely, we introduce a non-learning algorithm that requires no annotations to separate touching nuclei. Near-convex shape decomposition has been widely studied in its application to segment binary shapes into parts. However, little has been studied in its effectiveness in separating touch nuclei in histopathology images. We adapted a near-convex shape decomposition algorithm by developing novel ways of defining cuts and assigning pathologically reasonable weights to the cuts, which proved to be well suited for this task. The output of the separation algorithm is integrated into the compositional model.

We conducted experiments on an in-house DAPI (4',6-diamidino-2-phenylindole) stained histopathology image dataset. The empirical results demonstrate the effectiveness and data efficiency of the proposed method for nuclei detection and weakly-supervised segmentation.

In summary, our contributions are as follows:
\begin{itemize}
\item We proposed a method that gives probabilistic generative modeling of nuclei features, and makes predictions in a human interpretable way. 
\item We introduced a convex-shape decomposition algorithm to nuclei detection and segmentation, which is integrated as an important part into the probabilistic model.
\item We empirically showed that the proposed method surpassed a patch-based CNN network with the same field of view size. When trained with the same amount of data, the proposed method even surpassed a state-of-the-art fully-convolutional neural network that takes full-resolution supervision.
\end{itemize}

\section {Related Work}
\textbf{Nuclei Detection} 
There have been a surge of interest in the application of deep learning to nuclei detection~\cite{Wang2016,Xu2016,Kashif2016,Rojas-Moraleda2017}. \cite{Zhou2018} designed a Fully-Convolutional Network (FCN) with sibling branches that interactively handle nuclei detection and fine-grained classification. \cite{Schmidt2018} proposed to localize cell nuclei via star-convex polygons, which showed as a better shape representation than bounding boxes. Some works adapt a top-down object detector to histopathology images. \cite{Du2019} adapts a Faster-RCNN-FPN network with deformable convolution layers and context-aware module for the detection of abnormal cervical cells. \cite{Baykal2020} thoroughly studied the adaptation of Faster-RCNN~\cite{ren2015faster}, R-FCN~\cite{dai2016r} and SSD~\cite{liu2016ssd} for nuclei detection on pleural effusion cytology images. Another thread of work is to formalize detection as regression. \cite{Xie2015} proposed structured regression to a proximity map representation, where values on the proximity map represent the proximity to or probability of a nucleus center. Detection of cell centroids are obtained by finding the maximum positions on the proximity map. \cite{Sirinukunwattana2015} introduced spatially constrained layer to predict the probability of a pixel being the center of a nucleus. \cite{Hofener2018} showed that the prediction of a proximity map can be formed as either a classification or regression task. Most of these works use a fairly deep backbone (\eg~ResNet~\cite{he2016deep} or U-Net~\cite{Ronneberger2015}) and require annotations of all nuclei in the dataset. 

\textbf{Weakly-supervised Nuclei Segmentation}
Recently, there have been interests that explore nuclei segmentation with weak supervision. Most works exploited pseudo labels generated in various ways. \cite{Lee2020} generates pseudo-labels and progressively generates full training label with scribble annotations. \cite{Qu2020a} uses Voronoi algorithm and KMeans clusterning to derive coarse labels from nucleus point annotation. \cite{guo2021learning} first trained a network with boundary and super-pixel annotations as pseudo-labels, then applied confident learning to correct the pseudo-labels for a refined training. \cite{yoo2019pseudoedgenet} proposed an PseudoEdgeNet to extract nucleus edges without edge annotation as supervisory signals for the main segmenation network. Active learning is another way to reduce annotation cost. \cite{Qu2020} predicted uncertainty by a BayesianCNN then selected hard nuclei for manually mask annotation. These works saved labelling effort by using weak annotation. However, a large collection of data is still needed, and same as supervised learning, the decision process of deep neural networks are not interpretable.

\section{Method}
\label{sec:method}
In Section~\ref{sec:comp}, we discuss the application of Compositional Network~\cite{Kortylewski2020,kortylewski2020compositionalijcv}, which was originally introduced for natural image classification, to the modeling of nucleus image. We discuss its interpretability in Section~\ref{sec:interpretability} and our extension based on the Compositional Networks to the task of multiple instance detection in Section~\ref{sec:comp_multi}, where each nucleus is regarded as an object instance. Finally, in Section~\ref{sec:decomp}, we discuss the intergration of prior knowledge about the the near-convex shape of nuclei into the developed probabilistic compositional model, which further facilitates the separation of touching nuclei. The flowchart of the whole method is illustrated in Figure~\ref{fig:flowchart}.

\subsection{Compositional Networks for Nuclei Modeling}
\label{sec:comp}
Compositional Network~\cite{Kortylewski2020} explains the feature map from a convolutional layer in a generative view. Denote a feature map as $F \in \mathbb{R}^{H \times W \times D}$, with $H$ and $W$ being the spatial size and $D$ being the channel size. The feature vector $f_i$ at position $i$ are assumed independently generated, and each is modeled as a mixture of von-Mises-Fisher (vMF) distributions:

\begin{equation}
p(F|\mathcal{A}, \Lambda) = \prod_{i} p(f_i|\mathcal{A}_{i},\Lambda), \label{eq:vmf}
\end{equation}
\begin{equation}
p(f_i|\mathcal{A}_{i},\Lambda) = \sum_k \alpha_{i,k} p(f_i|\mu_k),\label{eq:vmf2}
\end{equation}
\begin{equation}
p(f_i|\mu_k) \propto \exp{\{\sigma f_i^T \mu_k\}}, \text{$|| f_i || = 1, || \mu_k || = 1$}.
\end{equation}

where $\Lambda=\{\mu_k\}$ are kernels for vMF distribution, which can be regarded as the ``mean" feature vector of each mixture component $k$, and $\mathcal{A}_{i}=\{\alpha_{i,k}\}$ are the spatial coefficients, which learn the probability of $\mu_k$ being activated at position $i$. We set the hyperparameter $\sigma$ in vMF distribution as fixed for tractability. Given a set of feature maps from a pre-trained CNN, the mixture coefficients $\{\alpha_{i,k}\}$ and the vMF kernels $\{\mu_k\}$ can be learned via Maximum Likelihood Estimation.

\subsection{Interpretable Modeling of Nucleus}
\label{sec:interpretability}
The vMF kernels represent image patterns that frequently occur in the training data. In prior work \cite{Kortylewski2020,kortylewski2020compositionalijcv} the kernels were shown to correspond to object parts, such as tires of a car. 
We observe a similar property, as the feature vectors that have high cosine similarity with the vMF kernels resemble certain nucleus parts (background, edges or interior patterns), see Section~\ref{sec:exp_det}. We say that a vMF kernel $\mu_k$ is activated at position $i$ if $f_i$ and $\mu_k$ have a high cosine similarity.

An important property of convolutional networks is that the spatial information from the image is preserved in the feature maps. To utilize this property, the set of spatial coefficients $\{\alpha_{i,k}\}$ are introduced to describe the expected activation of a kernel $\mu_k$ at a position $i$. Thus, $\alpha_k$ at all positions can be intuitively thought of as a 2D template, which depicts the expected spatial activation pattern of parts in an image of a nucleus -- \eg~where the edges are expected to be located in the image.
Therefore, the decision process of the proposed model can be interpreted as first detecting parts, then spatially combining them into a probability about the nucleus' presence. Note that this implements a part-based voting mechanism.

As the spatial pattern of parts varies dramatically with the shape, size and orientation of a nucleus, we further represent $F$ as a mixture of compositional models:
\begin{equation}
\label{eq:mix}
p(F|\Theta) = \sum_{m=1}^M \nu_m p(F|\mathcal{A}^m),
\end{equation}
with  $\mathcal{V}\texttt{=}\{\nu^m \in\{0,1\}, \sum_m \nu_m\texttt{=} 1 \}$. Here $M$ is the number of compositional models in the mixture distribution and $\nu_m$ is a binary assignment variable that indicates which compositional model is active.
Intuitively, each mixture component $m$ will represent a different set of nuclei with specific shape and size (see Section~\ref{sec:exp_det}). The parameters of the mixture components $\{\mathcal{A}^m\}$ need to be learned in an EM-style manner by iterating between estimating the assignment variables $\mathcal{V}$ and maximum likelihood estimation of $\{\mathcal{A}^m\}$ \cite{Kortylewski2020}.

\subsection{Adaptation to Nucleus Detection}
\label{sec:comp_multi}
Previous work has proposed to detect salient object in natural images based on Compositional Network~\cite{wang2020}. However, it is limited by the assumption that only one salient object is present in an image. Due to the significant difference between histopathology images and natural images, the adaptation for nucleus detection is non-trivial. 

First, the background in DAPI stained histopathology images is cleaner than natural images. However, this encourages the model to rely heavily on the background signals, which is undesirable and results in false positives in background regions.
We propose to get rid of the disturbance of background signals by masking. For each mixture component $m$, we pick a subset from $\mathcal{A}^m$ to obtain a soft foreground mask:
\begin{align}
    M^m = \sum_{k \in K_{f}} \alpha^m_{k}
\end{align}
where $K_f$ is a subset of vMF kernels which represents foreground parts (interior, edge, etc.). Then, we modify the computation of log-likelihood of $p(F|\mathcal{A}^m)$ as:
\begin{align}
    \log p(F|\mathcal{A}^m) = \frac{\sum_i M^m_i \log p(f_i|\mathcal{A}^m_i, \Lambda)}{\sum_i M^m_i}
\end{align}
which gives more weights to vMF kernels activated at foreground.

Second, we extend the model to multiple objects by modifying the likelihood:
\begin{equation}
\label{eq:multi}
    p(\mathcal{F}) = \prod_i \prod_n p(F_i)^{z_{i,n}}
\end{equation}
where $F_i$ are patches from a whole feature map $\mathcal{F}$, and $\{z_{i,n} \in \{0,1\}|\sum_n z_{i,n} = 1\}$ are indicators of existence of object $n$ at patch $F_i$. Note that by the design of the likelihood, only one object model can be active at one position in the feature map. We maximize the likelihood defined in Equation~\ref{eq:multi} by applying the model in sliding windows, then selecting the local maxima in the resulted likelihood map after non-maximum suppression.

\subsection{Touching Nuclei Separation}
\label{sec:decomp}
Nuclei usually clump and touch with each other, which makes it difficult to recognize single nucleus. The compositional model is able to explain for a single nucleus, but insufficient to separate touching nuclei precisely. We introduce a non-learning-based algorithm to segment nucleus that requires no annotations. The algorithm is adapted from near-convex shape decomposition~\cite{Ren2011}, and is integrated into the compositional model as a prior, telling the model where to pay attentions. 

First, we select the vMF kernel $\mu_{0}$ that respond to the background. Given a feature map $F$, we compute $1 - \mu_0^T f_i$ at each position $i$ to obtain a nucleus foreground score map. A threshould is applied on the score map to give foreground connected components. These connected components may consist of a single nucleus or touching nuclei. To distinguish between them, we leverage the following observations: 1) The shapes of nuclei are usually convex. 2) When multiple nuclei cluster together, there are usually concave points along the boundary of the connected component. Based on these observations, we propose to use a near-convex shape decomposition algorithm to process each connected component.

\begin{figure}[h]
\vskip 0.2in
    \centering
    \includegraphics[width=0.4\textwidth]{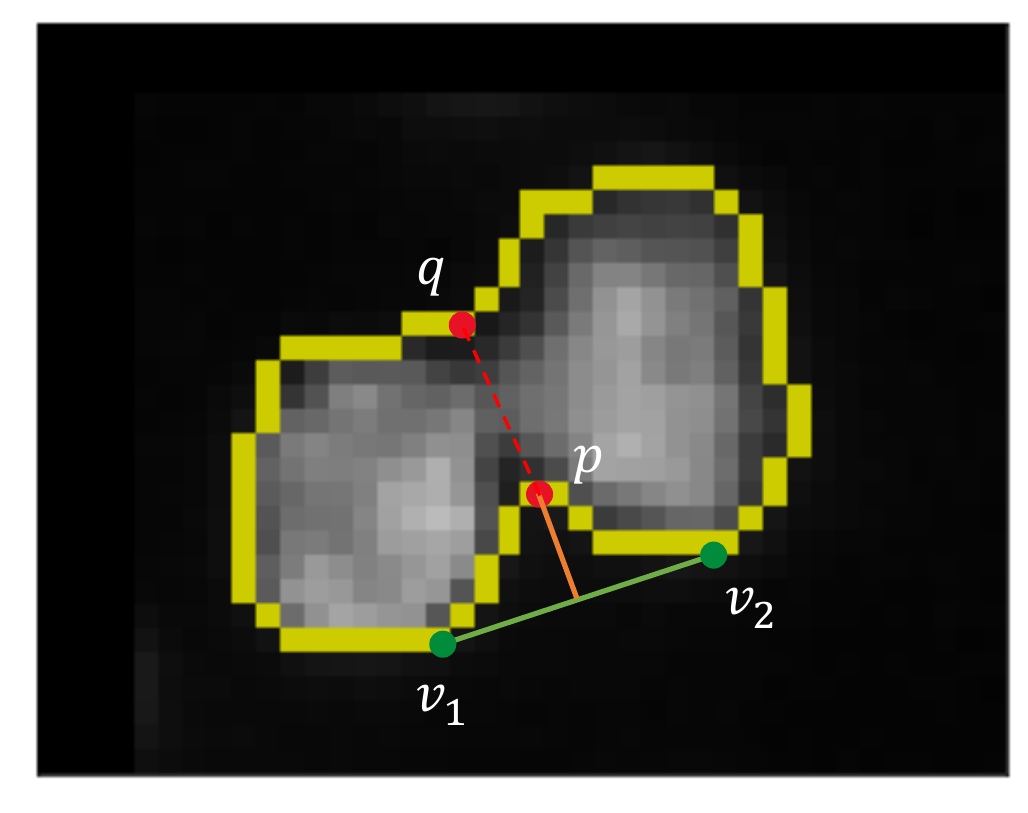}
    \caption{An example of touching nuclei. The boundary is colored in yellow. A concave point $p$ on the boundary is labelled in red. The line segment between two points $v_1$, $v_2$ on the boundary are outside of the nuclei, and $c(v_1, v_2)$ is given by the distance from $p$ to $v_1 v_2$. A cut $pq$ is needed to separate $v_1$, $v_2$ in this case.}
    \label{fig:touching}
\vskip 0.1in
\end{figure}

Following \cite{Ren2011}, a near-convex decomposition of a shape $S$, $D_{\phi}(S)$, is defined as a set of non-overlapping parts:
\begin{align}
    D_{\psi}(S) & = \{P_i | \bigcup_i P_i = S, \forall P_i \cap P_j = \emptyset, c(P_i) \le \psi \} \\
    c(P_i) & = \max_{v_1, v_2 \in Boundary(P_i)}\{c(v_1, v_2)\} \\
    c(v_1, v_2) & =
    \begin{cases}
        \max_{u} d(u, v_1v_2)  & \text{if $v_1 v_2$ is outside $P_i$} \\
        0 & \text{otherwise}
    \end{cases}
\end{align}
where $P_i$ denotes the decomposed parts; $\psi$ is a parameter for near-convex tolerance. Concavity $c(P_i)$ is given by max of concavity $c(v_1, v_2)$ for any two points $v_1, v_2$ on the boundary of $P_i$, which is intuitively defined as the max distance from a point $u$ between $v_1, v_2$ on the boundary to the line segment $v_1v_2$. Besides, a set of potential cuts is needed to split $S$. We compute the curvature of the boundary of $S$, and find concave points with negative curvature. A potential cut is formed by the line segment between two concave points if the line segment lies inside $S$. To comply with the near-convex constraint, all $v_1, v_2$ with $c(v_1, v_2) > \psi$ must be cut into different parts. These pairs of points are named mutex pairs. An illustration is given in Figure~\ref{fig:touching}.

Let $N$ be the number of potential cuts and $M$ be the number of mutex pairs. We record two matrices $A_{N\times M}$ and $B_{N\times N}$, where $A$ indicates whether a cut splits a mutex pair, and $B$ indicates whether two cuts intersect with each other:
\[ A_{ij} =
\begin{cases}
1 & \text{if cut i splits mutex pair j} ,\\
0 & \text{otherwise} .
\end{cases} \]
\[
 B_{ij} = 
\begin{cases}
1 & \text{if cut i intersects with cut j}, \\
0 & \text{otherwise}.
\end{cases}\]
Following \cite{Ren2011}, the shape decomposition is formed as an integer programming problem:
\begin{equation}
    \min w^T x \ \ \ s.t. Ax \ge 1, x^TBx = 0, x \in \{0, 1\}^N
\end{equation}
where $x$ is a Boolean selection variable for each potential cut. The first constraint says that every mutex pair should be splitted by some cut, and the second one says that the selected cuts should not intersect with each other. $w$ is a weight vector for each cut 
defined as:
\begin{equation}
    w_{pq} = \exp\{\langle \vec{n_p}, \vec{pq} \rangle\} + \exp\{\langle \vec{n_q}, \vec{qp} \rangle\} +  \exp\{\lambda ||\vec{pq}||\},
\end{equation}
where $p$ and $q$ are cut endpoints, $\vec{n}$ is the normal vector of the boundary, and $\langle \cdot, \cdot \rangle$ denotes the angle between two vectors in degree. $\lambda$ is set to $0.1$. The defined cut weights encourage the selected cuts to be perpendicular to the local boundary and be short, which are in accord with human intuition.

\begin{figure*}[ht]
\vskip 0.2in
\centering
\includegraphics[width=0.9\textwidth]{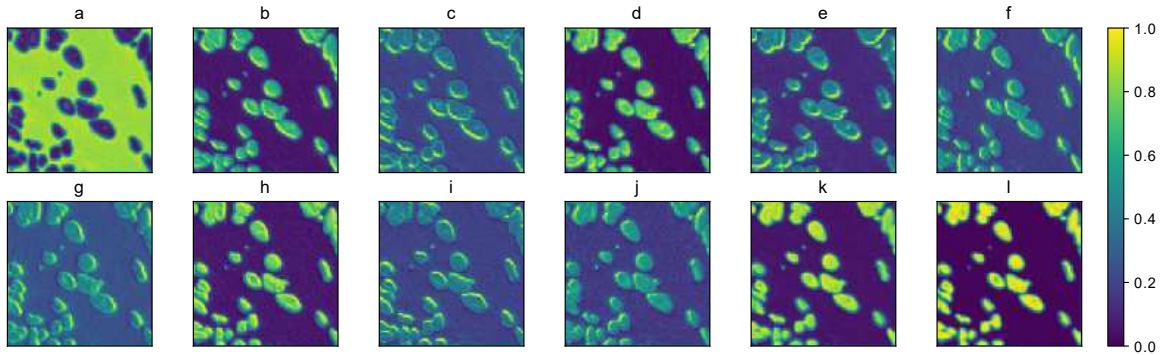}
\caption{Activations of vMF kernels on a sample image. (a) is a kernel activated at background, (b-j) are kernels mostly activated near edges, and (k-l) are kernels activated by the nuclei interior texture.}
\label{fig:vc}
\vskip 0.1in
\end{figure*}

\subsubsection{Nuclei Candidates as Prior} 
After decomposing the nuclei foreground connected components, the obtained regions are near convex and are taken as candidates of single nucleus. These candidates serve as a prior guiding where to pay attention to for nucleus detection. We define the prior probability of nucleus existence $q$ as Gaussian distributions centered at each candidate. The final detection probability is obtained by integrating the prior into the compositional model and the final probability map is defined as:
\begin{equation}
    p(\mathcal{F}) = \prod_i \prod_n p(F_i)^{z_{i,n}}q(i).
\end{equation}

\subsection{Weakly-supervised Nuclei Segmentation}
\label{sec:seg}
The nuclei candidates obtained from Section~\ref{sec:decomp} can also be used as segmentation masks. Since the algorithm only receives bounding box as supervision which is used to crop nucleus images, it achieves segmentation masks in a weakly-supervised way. The obtained segmentation masks have a property to be near convex. Although rare nuclei can have concave shapes and be wrongly cut, it can be indicative of potential annotation errors (\eg~where the annotator mistakenly recognized a pair of touching nuclei as a single one). 


\section{Experiments and Results}
\textbf{Dataset.}
Even though hematoxylin and eosin (H\&E) stain is the most widely used stain and often the gold standard to visualize the cell structures of histological section, there have been efforts to develop different types of stains and image modalities. Especially, multiplexed immunofluorescence (mIF) and immunohistochemistry (IHC) are emerging technologies with better predictions \cite{Lu2019} for immunotherapy.

The mIF images were obtained using Vectra-3 and Vectra Polaris microscopes (Akoya BioSciences, MA, USA) from six patients with liver cancer (3), lung adenocarcinoma (1), lung small cell carcinoma (1), and melanoma (1). The whole slide consists of hundreds of fields in general, and each field has $1872 \times 1404$ ($width \times height$) size with $32$-bit pixel-depth. Total $6$ fields, randomly selected from each patient, were used for the experiment in this study. For the nuclei detection and segmentation, DAPI stained images were used in this study among the multispectral images. 
The selected fields were manually checked and annotated by trained researchers and totally 18312 nuclei were annotated. Each field was divided to ($256 \times 256$) patches which gives totally 210 patches, from which $186$ patches were used for training and $24$ patches were used for testing.


\subsection{Nuclei Detection}
\label{sec:exp_det}
\textbf{Evaluation metrics.}
Following~\cite{Sirinukunwattana2015,Kashif2016,Alom2018,Hagos2019}, we adopt the commonly used precision (P) - recall (R) metrics to evaluate nucleus detection methods. 
If a predicted nucleus central point is in the proximity of 3 pixels to a ground truth point, it is regarded as true positive, otherwise it is regarded as false positive. If a ground truth point has no corresponding prediction, it is regarded as false negative. By binarizing the predicted score map at different thresholds, we can get a series of precision and recall values and plot them as P-R curve.

\textbf{Implementation details.}
The architecture of the proposed method consists of the first layer of U-Net \cite{Ronneberger2015} as a feature extractor. This layer is pre-trained along with the U-Net by super-pixel segmentation as a pretext task. It is followed by a generative compositional model as defined in Section~\ref{sec:method}. The proposed design leads to a light-weight model, which has totally only 213K parameters.

\begin{figure*}[ht]
\vskip 0.2in
\centering
\includegraphics[width=0.9\textwidth]{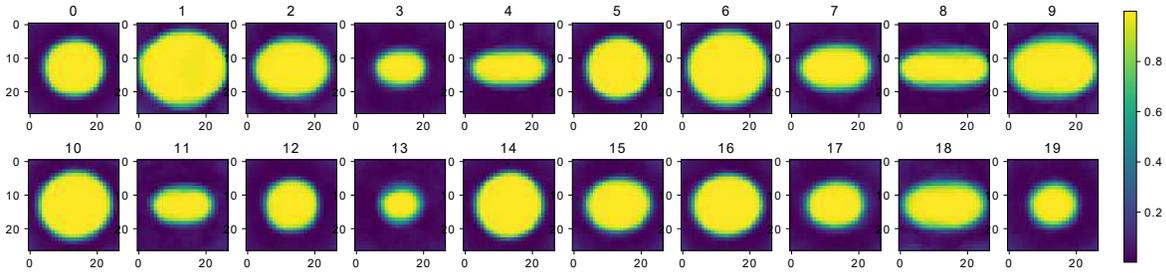}
\caption{Average nucleus foreground pattern in each mixture component. Different mixture components are responding to nuclei with different sizes and shapes.} \label{fig:mix}
\vskip -0.2in
\end{figure*}

The vMF kernels (Figure~\ref{fig:vc}) are learned in an unsupervised manner by maximum likelihood estimation. We empirically found that 12 kernels are sufficient to model parts of a nucleus. Among them one is activated mostly at background, nine of them are activated mostly near edges of nuclei, and the last two have highest response to the interior regions of nuclei.
We extracted 3097 isolated nuclei to learn the mixture parameters $\{\mathcal{A}^m\}$. These nuclei were cropped as 27x27 patches, aligned by orientation, and clustered into $M=20$ clusters using KMeans by their length of long axis and short axis. A compositional model was learned for each cluster of nuclei with specific size and shape (Figure~\ref{fig:mix}). To detect nuclei with various orientations, we rotated images by [-90, -60, -30, 30, 60] degrees before input, and restore the original rotation after getting output from the model.
To improve the performance on touching nuclei, we apply the near-convex decomposition prior as discussed in Section \ref{sec:decomp}, where $\psi$ was set to 3 pixels and the variance of the Gaussian distribution was $\sigma=10$.

\begin{figure}[h]
\vskip 0.2in
    \centering
    \includegraphics[width=0.45\textwidth]{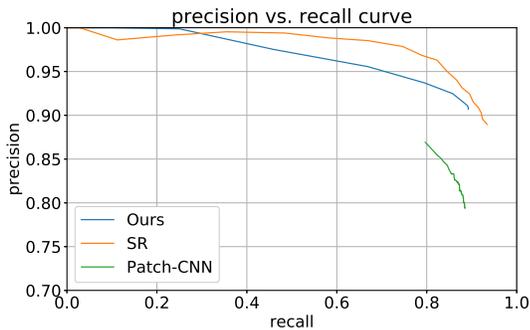}
    \caption{Evaluation of nuclei detection by Precision-Recall curve.}
    \label{fig:result}
\vskip -0.1in
\end{figure}

\begin{figure}[h]
\vskip 0.2in
    \centering
    \includegraphics[width=0.45\textwidth]{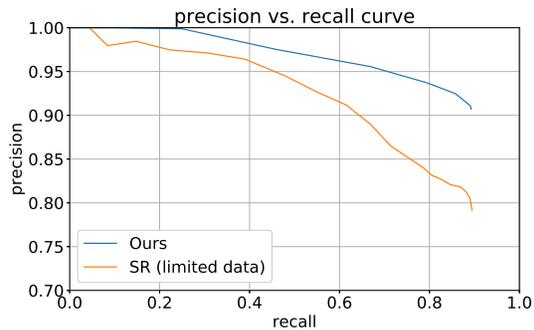}
    \caption{Comparison of the proposed method and SR model with approximately same amount of training samples.}
    \label{fig:ablation}
\vskip 0.1in
\end{figure}

\textbf{Baselines.} Recently many advanced deep learning models have achieved state-of-the-art performance on nuclei detection~\cite{Hofener2018}. However, our motivation is to develop data-efficient models in order to save annotation efforts, at the same time exploit internal representations to make it interpretable. Therefore, We compare our model with a classic baseline: a patch-CNN~\cite{Sirinukunwattana2016c} and one of the state-of-the-art methods~\cite{Xie2018} which utilized structured regression with a U-Net-like backbone~\cite{Ronneberger2015}. The CNN model is designed for nucleus detection and comprises of 2 convolution layers with maxpooling and 3 fully connected layers. We adapted the last spatially constrained layer to a softmax layer to classify image patchs. If an image patch locates within 3 pixels to a nucleus centroid, it is labelled positive, otherwise negative. During testing, the network predicts the probability of being a nucleus centroid for each pixel. Same patch size was used in the proposed method, which makes the comparison fair. Note that this network has 893K parameters, 3 times more than the proposed model.

The structured regression (SR) method~\cite{Xie2018} is trained with full image supervision, as it was designed. Therefore, it serves as an upper-bound for comparison, since the proposed method only used image patches of isolated cells, which is a subset of the entire images. It is trained for 20 epochs by Adam optimizer with learning rate 1e-4 and batch size 8. During inference, local maxima on the regressed map with non-maximum suppression are obtained as detected nucleus centers. 

\begin{table}
\centering
\caption{Weakly supervised segmentation performance measured in AJI and DSC on the in-house dataset.} \label{tab:seg}
\vskip 0.2in
\begin{tabular}{|p{0.2\textwidth}|p{0.1\textwidth}|p{0.1\textwidth}|}
\hline
                   & AJI    & DSC\\
\hline
BBTP               & 0.6765   & 0.8513         \\
PointAnno          & 0.5991   & 0.7805         \\
Ours               & \textbf{0.7030}   & \textbf{0.8900} \\
\hline
\end{tabular}
\vskip -0.2in
\end{table}

\begin{figure*}[ht]
\centering
\includegraphics[width=0.8\textwidth]{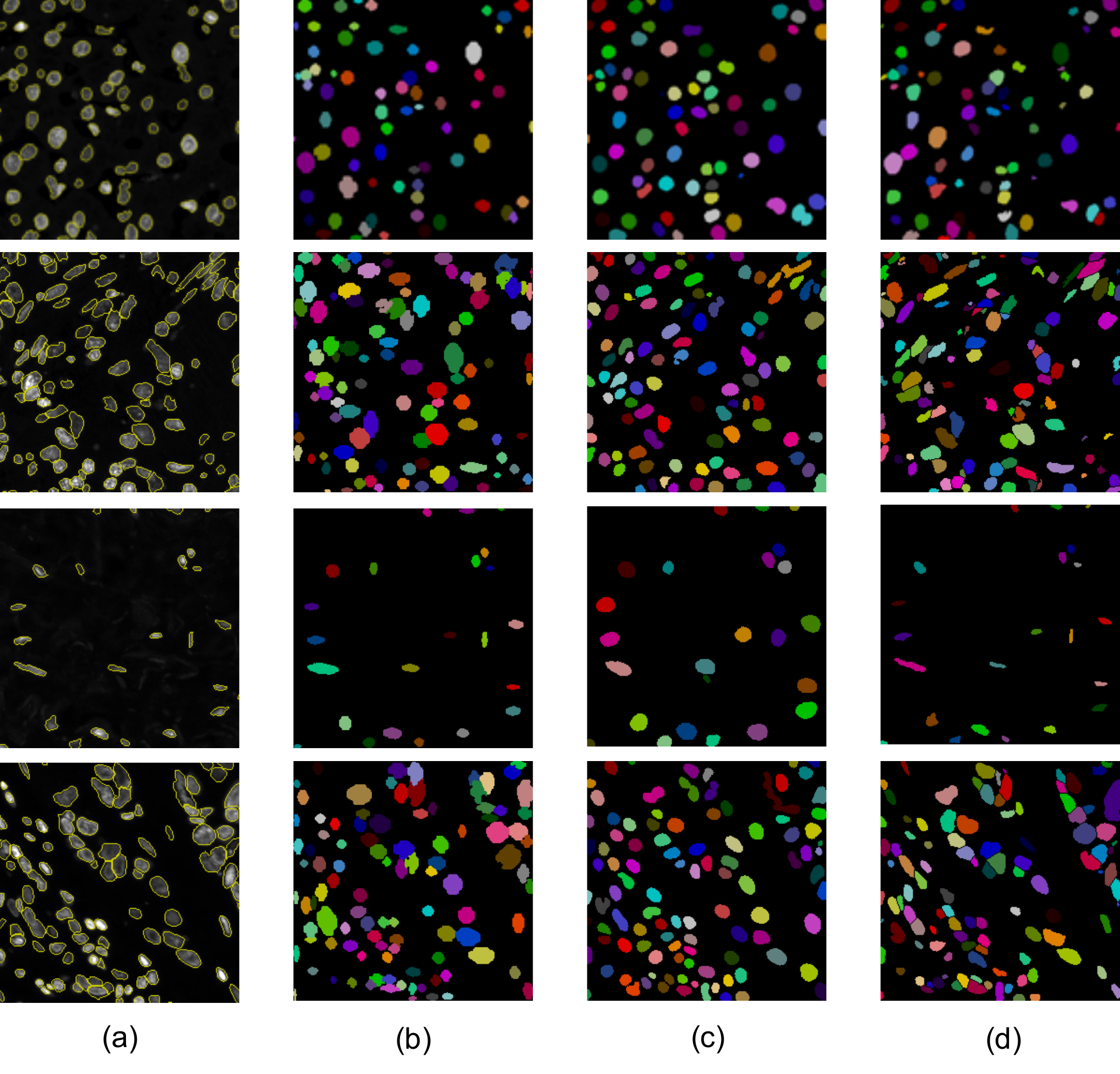}
\caption{Qualitative segmentation results. Each line shows one image from the test set and the segmentation masks obtained by three methods on it. (a) is the DAPI stained image with the boundaries of each nucleus annotated. (b) are the predictions of BBTP, (c) are the predictions of PointAnno, and (d) are the predictions of the proposed method.}
\label{fig:qual}
\vskip 0.1in
\end{figure*}

\textbf{Results.} Figure~\ref{fig:result} shows the Precision-Recall curve of the baseline Patch-CNN, SR and the proposed method. The Patch-CNN has exactly the same size of field of view with the proposed model, while has more number of parameters than the latter. However, the proposed method surpass it by a large margin, which shows the effectiveness of the proposed method. We believe this is due to the explicit generative modeling of nuclei features, which boosts performance while keeping the model to be light-weight.

Due to extended model complexity and full image resolution training, SR outperforms the proposed method. This is understandable since the SR model is a much deeper network with more than 20 convolutional layers, and has a larger field of view with full image supervision.
However, deep neural networks like SR are data hungry and need large amounts of data to learn their parameters, while the proposed method only requires the annotations of isolated nuclei, which saves much effort for human experts compared with annotating the entire image.

To verify the hypothesis that the proposed method is more data-efficient than deep neural networks, we made a comparison between SR and the proposed method under approximately the same amount of training data in terms of the number of nuclei used. In Figure~\ref{fig:ablation}, we can see that when trained with approximately the same amount of data, the performance of SR degrades significantly. This result proves that when large amounts of annotated nuclei samples are not available, the proposed method is able to present superior nuclei detection results than an over-parameterized (in terms of the dataset size) deep neural network.

\subsection{Weakly-supervised Nuclei Segmentation}
The learning of the proposed method only requires the annotation of nucleus positions and bounding boxes. As stated in Section~\ref{sec:seg}, by utilizing the unsupervisedly learned vMF kernels and the near-convex decomposition algorithm, we can obtain nuclei instance segmentation masks. We compare with two weakly-supervised segmentation methods, BBTP \cite{Hsu2019} and PointAnno \cite{Qu2020a}. BBTP is a well-known weakly-supervised model for natural images, and PointAnno is developed for nuclei segmentation. Aggregated Jaccard Index (AJI) \cite{Kumar2017} and Dice similarity coefficient (DSC) were used as metrics:
\begin{equation}
    AJI=\frac{\sum_{i=1}^{n} |G_i \cap S(G_i)|}{\sum_{i=1}^n |G_i \cup S(G_i)| + \sum_{k \in K} |S_k|},
\end{equation}
\begin{equation}
    DSC = \frac{2 |A \cap B|}{|A| + |B|}
\end{equation}
where $S(G_i)$ is the segmented object that has maximum overlap with $G_i$, $K$ is the set of segmented objects that are not assigned to ground-truth objects, and $A$ and $B$ are pixels of ground-truth foreground and segmentation foreground respectively. AJI focuses more on the correct matching between segmented nuclei instances and ground-truths, while DSC focuses on the foreground/background classification. 

Table~\ref{tab:seg} shows the segmentation performance of the three methods. The proposed method outperforms BBTP and PointAnno, which verified the effectiveness of our method. What's more, the proposed method requires little training (only the clustering of vMF kernels), which is an advantage over deep networks. Qualitative results are shown in Figure~\ref{fig:qual}. The proposed method is able to precisely locate the foreground and cut touching nuclei, even for hard cases where more than two nuclei are touching with each other. Compared with BBTP and PointAnno, the segmentation masks obtained by the proposed method aligns better with the ground-truth nuclei contours, thanks to the accurate detection of foreground as well as the cutting at reasonable positions between touching nuclei.

\section{Conclusion}
We introduced a light-weight interpretable model to the task of nuclei detection and segmentation for DAPI stained histopathology images. It has several merits. First, it is data-efficient, in that it is able learn only from annotations of isolated nuclei and significantly reduce the annotation efforts; Second, it is light-weight, in that it is composed of only one convolutional layer, one layer of vMF kernels and one layer of spatial coefficients; Third, it is interpretable. Thanks to exploitation of the patterns of nuclei and the explicit modeling, the model is able to give its prediction with sound explanation in probability. We also introduced a convex-shape decomposition algorithm to nuclei detection and segmentation, which is integrated as in important part into the probability model. Empirical results proved that the proposed method is able to achieve satisfactory performance on the task of nucleus detection and segmentation, especially when the number of available annotations are limited. For future study, we expect that the proposed method can be combined with transfer learning, where features pre-trained on one histopathology dataset can be exploited to build models on another dataset with different stains.


\nocite{langley00}

\bibliography{refs}
\bibliographystyle{icml2022}



\end{document}